\documentclass{article}

\usepackage[final]{arxiv}

\usepackage[utf8]{inputenc} 
\usepackage[T1]{fontenc}    
\usepackage{url}            
\usepackage{booktabs}       
\usepackage{amsfonts}       
\usepackage{nicefrac}       
\usepackage{microtype}      

\usepackage[acronym]{glossaries}
\usepackage{amsmath, amsfonts, amsthm, amssymb}
\usepackage{subfig} 
\usepackage{multicol}

\title{Deep Reinforcement Learning from Self-Play in Imperfect-Information Games}

\author{
  Johannes Heinrich \\
  University College London, UK\\
  \texttt{j.heinrich@cs.ucl.ac.uk} \\
  \And
  David Silver \\
  University College London, UK \\
  \texttt{d.silver@cs.ucl.ac.uk} \\
}

\usepackage{graphicx} 
\usepackage{varwidth}

\usepackage{algorithm,algpseudocode}
\algnewcommand\algorithmicforeach{\textbf{for each}}
\algdef{S}[FOR]{ForEach}[1]{\algorithmicforeach\ #1\ \algorithmicdo}

\theoremstyle{definition}

\newcommand{\cond}{\,\vert\,}
\newcommand{\Cond}{\,\middle\vert\,}

\newacronym{mdp}{MDP}{Markov decision process}
\newacronym{pomdp}{POMDP}{partially observable Markov decision process}
\newacronym{fsp}{FSP}{Fictitious Self-Play}
\newacronym{fqi}{FQI}{Fitted Q Iteration}
\newacronym{nfq}{NFQ}{Neural Fitted Q Iteration}
\newacronym{cfr}{CFR}{Counterfactual Regret Minimization}
\newacronym{xfp}{XFP}{Full-Width Extensive-Form Fictitious Play}
\newacronym{nfsp}{NFSP}{Neural Fictitious Self-Play}
\newacronym{dqn}{DQN}{Deep Q Network}
\newacronym{sgd}{SGD}{Stochastic Gradient Descent}
\newacronym{acpc}{ACPC}{Annual Computer Poker Competition}
\newacronym{lhe}{LHE}{Limit Texas Hold'em}

\begin{document}

\maketitle

\begin{abstract}
Many real-world applications can be described as large-scale games of imperfect information. 
To deal with these challenging domains, prior work has focused on computing Nash equilibria in a handcrafted abstraction of the domain. 
In this paper we introduce the first scalable end-to-end approach to learning approximate Nash equilibria without prior domain knowledge. 
Our method combines fictitious self-play with deep reinforcement learning. 
When applied to Leduc poker, \gls{nfsp} approached a Nash equilibrium, whereas common reinforcement learning methods diverged. 
In Limit Texas Hold'em, a poker game of real-world scale, \gls{nfsp} learnt a strategy that approached the performance of state-of-the-art, superhuman algorithms based on significant domain expertise.

\end{abstract} 

\section{Introduction}
\label{introduction}

Games have a tradition of encouraging advances in artificial intelligence and machine learning 
\citep{samuel1959some,tesauro1995temporal,campbell2002deep,gelly2012grand,bowling2015heads}. A game is a domain of conflict or cooperation between several entities \citep{myerson1991game}. 
One motivation for studying recreational games is to develop algorithms that will scale to more complex, real-world games such as airport and network security, financial and energy trading, traffic control and routing \citep{lambert2005fictitious,nevmyvaka2006reinforcement,bazzan2009opportunities,tambe2011security,urieli2014tactex}. 
Most of these real-world games involve decision making with imperfect information and high-dimensional information state spaces. 
An optimal (and in principle learnable) solution to these games would be a Nash equilibrium, i.e. a strategy from which no agent would choose to deviate. While many machine learning methods have achieved near-optimal solutions to classical, perfect-information games, these methods fail to converge in imperfect-information games.
On the other hand, many game-theoretic approaches for finding Nash equilibria lack the ability to learn 
abstract patterns and use them to generalise to novel situations.
This results in limited scalability to large games, unless the domain is abstracted to a manageable size using human expert knowledge, heuristics or modelling. 
However, acquiring human expertise often requires expensive resources and time. In addition, humans can be easily fooled into irrational decisions or assumptions \citep{selten1990bounded}. 
This motivates algorithms that learn useful strategies end-to-end.

Fictitious play \citep{brown1951iterative} is a popular method for learning Nash equilibria in normal-form (single-step) games. Fictitious players choose best responses (i.e. optimal counter-strategies) to their opponents' average behaviour. \gls{fsp} \citep{Heinrich15FSP} extends this method to extensive-form (multi-step) games.
In this paper we introduce \gls{nfsp}, a deep reinforcement learning method for learning \emph{approximate} Nash equilibria of imperfect-information games. \gls{nfsp} combines \gls{fsp} with neural network function approximation. 
An \gls{nfsp} agent consists of two neural networks. The first network is trained by reinforcement learning from memorized experience of play against fellow agents. This network learns an approximate best response to the historical behaviour of other agents. The second network is trained by supervised learning from memorized experience of the agent's own behaviour. This network learns a model that averages over the agent's own historical strategies. The agent behaves according to a mixture of its average strategy and best response strategy.

We empirically evaluate our method in two-player zero-sum computer poker games. In this domain, current game-theoretic approaches use heuristics of card strength to abstract the game to a tractable size \citep{johanson2013evaluating}. 
Our approach does not rely on engineering such abstractions or any other prior domain knowledge. \gls{nfsp} agents leverage deep reinforcement learning to learn directly from their experience of interacting with other agents in the game.
When applied to Leduc poker, \gls{nfsp} approached a Nash equilibrium, whereas common reinforcement learning methods diverged. 
We also applied \gls{nfsp} to \gls{lhe}, learning directly from the raw inputs. \gls{nfsp} 
learnt a strategy that approached the performance of state-of-the-art, superhuman methods based on handcrafted abstractions.

\section{Background} 

In this section we provide 
a brief overview of reinforcement learning, extensive-form games and fictitious self-play.
For a more detailed exposition we refer the reader to \citep{sutton1998reinforcement}, \citep{myerson1991game}, \citep{fudenberg1998theory}  and \citep{Heinrich15FSP}.

\subsection{Reinforcement Learning}

Reinforcement learning \citep{sutton1998reinforcement} agents typically learn to maximize their expected future rewards from interaction with an environment. The environment is usually modelled as a \textbf{\gls{mdp}}. An agent behaves according to a \textbf{policy} that specifies a distribution over available actions at each state of the \gls{mdp}. The agent's goal is to improve its policy in order to maximize its \textbf{gain}, $G_t=\sum_{i=t}^TR_{i+1}$, which is a random variable of the agent's cumulative future rewards starting from time $t$.

Many reinforcement learning algorithms learn from sequential \textbf{experience} in the form of transition tuples, $(s_t, a_t, r_{t+1}, s_{t+1})$, where $s_t$ is the state at time $t$, $a_t$ is the action chosen in that state, $r_{t+1}$ the reward received thereafter and $s_{t+1}$ the next state that the agent transitioned to. 
A common objective is to learn the \textbf{action-value function}, $Q(s,a)=\mathbb{E}^{\pi}\left[ G_t \Cond S_t = s, A_t = a \right]$, defined as the expected gain of taking action $a$ in state $s$ and following policy $\pi$ thereafter. 
An agent is learning \textbf{on-policy} if it learns about the policy that it is currently following. In the \textbf{off-policy} setting an agent learns from experience of another agent or another policy, e.g. a previous policy.

Q-learning \citep{watkins1992q} is a popular off-policy reinforcement learning method. It learns about the greedy policy, which at each state takes the action of the highest estimated value.
Storing and replaying past experience with off-policy reinforcement learning from the respective transitions is known as experience replay \citep{lin1992self}.
\gls{fqi} \citep{ernst2005tree} is a batch reinforcement learning method that replays experience with Q-learning.
\gls{nfq} \citep{riedmiller2005neural} and \gls{dqn} \citep{mnih2015human} are extensions of \gls{fqi} that use neural network function approximation with batch and online updates respectively.

\subsection{Extensive-Form Games}

\textbf{Extensive-form games} are a model of sequential interaction involving multiple players. Assuming rationality, each player's goal is to maximize his payoff in the game.
In imperfect-information games, each player only observes his respective \textbf{information states}, e.g. in a poker game a player only knows his own private cards but not those of other players. 
Each player chooses a \textbf{behavioural strategy} that maps information states to probability distributions over available actions. 
We assume games with \textbf{perfect recall}, i.e. each player's current information state $s_t^i$ implies knowledge of the sequence of his information states and actions, $s_1^i, a_1^i, s_2^i, a_2^i, ..., s_t^i$, that led to this information state.
The \textbf{realization-probability} \citep{von1996efficient}, $x_{\pi^i}(s_t^i) = \prod_{k=1}^{t-1} \pi^i(s_k^i,a_k^i)$, determines the probability that player $i$'s behavioural strategy, $\pi^i$, contributes to realizing his information state $s_t^i$.
A \textbf{strategy profile} $\pi = (\pi^1, ...\,  , \pi^n)$ is a collection of strategies for all players.
$\pi^{-i}$ refers to all strategies in $\pi$ except $\pi^i$.
Given a fixed strategy profile $\pi^{-i}$, any strategy of player $i$ that achieves optimal payoff performance against $\pi^{-i}$ is a \textbf{best response}. An approximate or $\epsilon$-best response is suboptimal by no more than $\epsilon$. 
A \textbf{Nash equilibrium} is a strategy profile such that each player's strategy in this profile is a best response to the other strategies. Similarly, an approximate or $\epsilon$-Nash equilibrium is a profile of $\epsilon$-best responses. In a Nash equilibrium no player can gain by deviating from his strategy. Therefore, a Nash equilibrium can be regarded as a fixed point of rational self-play learning. In fact, Nash equilibria are the only strategy profiles that rational agents can hope to converge on in self-play \citep{bowling2001rational}.

\subsection{Fictitious Self-Play}

\textbf{Fictitious play} \citep{brown1951iterative} is a game-theoretic model of learning from self-play. Fictitious players choose best responses to their opponents' average behaviour.
The average strategies of fictitious players converge to Nash equilibria in certain classes of games, e.g. two-player zero-sum and many-player potential games \citep{robinson1951iterative, monderer1996fictitious}.
\citet{leslie2006generalised} introduced generalised weakened fictitious play. It has similar convergence guarantees as common fictitious play, but allows for approximate best responses and perturbed average strategy updates, making it particularly suitable for machine learning.

Fictitious play is commonly defined in normal form, which is exponentially less efficient for extensive-form games. \citet{Heinrich15FSP} introduce \textbf{\gls{xfp}} that enables fictitious players to update their strategies in 
behavioural, extensive form, resulting in linear time and space complexity.
A key insight is that for a convex combination of normal-form strategies, 
$\hat{\sigma} = \lambda_1\hat{\pi}_1 + \lambda_2 \hat{\pi}_2$,
we can achieve a realization-equivalent behavioural strategy $\sigma$, by setting it to be proportional to the respective convex combination of realization-probabilities, 
\begin{equation}
\sigma(s,a) \propto \lambda_1 x_{\pi_1}(s)\pi_1(s,a) + \lambda_2 x_{\pi_2}(s)\pi_2(s,a) \quad \forall s,a,
\label{eq:real_equiv}
\end{equation}
where $\lambda_1 x_{\pi_1}(s) + \lambda_2 x_{\pi_2}(s)$ is the normalizing constant for the strategy at information state $s$. In addition to defining a full-width average strategy update of fictitious players in behavioural strategies, equation \eqref{eq:real_equiv} prescribes a way to sample data sets of such convex combinations of strategies.
\citet{Heinrich15FSP} introduce \textbf{Fictitious Self-Play (\gls{fsp})}, a sample- and machine learning-based class of algorithms that approximate \gls{xfp}. 
\gls{fsp} replaces the best response computation and the average strategy updates with reinforcement and supervised learning respectively. In particular,
\gls{fsp} agents generate datasets of their experience in self-play. 
Each agent stores its experienced transition tuples, $(s_t, a_t, r_{t+1}, s_{t+1})$, in a memory, $\mathcal{M}_{RL}$, designated for reinforcement learning. 
Experience of the agent's own behaviour, $(s_t, a_t)$, is stored in a separate memory, $\mathcal{M}_{SL}$, designated for supervised learning.
Self-play sampling is set up in a way that an agent's reinforcement learning memory approximates data of an \gls{mdp} defined by the other players' average strategy profile. Thus, an approximate solution of the \gls{mdp} by reinforcement learning yields an approximate best response. Similarly, an agent's supervised learning memory approximates data of the agent's own average strategy, which can be learned by supervised classification.

\section{Neural Fictitious Self-Play}

\gls{nfsp} combines \gls{fsp} with neural network function approximation. In algorithm \ref{algo:nfsp} all players of the game are controlled by separate \gls{nfsp} agents that learn from simultaneous play against each other, i.e. self-play.  An \gls{nfsp} agent interacts with its fellow agents and memorizes its experience of game transitions and its own best response behaviour in two memories, $\mathcal{M}_{RL}$ and $\mathcal{M}_{SL}$.
\gls{nfsp} treats these memories as two distinct datasets suitable for deep reinforcement learning and supervised classification respectively. The agent trains a neural network, $Q(s,a\cond\theta^Q)$, to predict action values from data in $\mathcal{M}_{RL}$ using off-policy reinforcement learning. 
The resulting network defines the agent's approximate best response strategy, $\beta=\epsilon\textnormal{-greedy}(Q)$, which selects a random action with probability $\epsilon$ and otherwise chooses the action that maximizes the predicted action values.
The agent trains a separate neural network, $\Pi(s,a\cond\theta^\Pi)$, to imitate its own past best response behaviour using supervised classification on the data in $\mathcal{M}_{SL}$. This network maps states to action probabilities and defines the agent's average strategy, $\pi=\Pi$.
During play, the agent chooses its actions from a mixture of its two strategies, $\beta$ and $\pi$.
\gls{nfsp} also makes use of two technical innovations in order to ensure the stability of the resulting algorithm as well as enable simultaneous self-play learning. First, it uses reservoir sampling \citep{vitter1985random} to avoid windowing artifacts due to sampling from a finite memory. Second, it uses anticipatory dynamics \citep{shamma2005dynamic} to enable each agent to both sample its own best response behaviour and more effectively track changes to opponents' behaviour.

\begin{algorithm}
    \caption{Neural Fictitious Self-Play (\gls{nfsp}) with fitted Q-learning}
    \label{algo:nfsp}
    \begin{algorithmic}
    \State Initialize game $\Gamma$ and execute an agent via \Call{RunAgent}{} for each player in the game
    \Function{RunAgent}{$\Gamma$}
    	\State Initialize replay memories $\mathcal{M}_{RL}$ (circular buffer) and $\mathcal{M}_{SL}$ (reservoir)
    	\State Initialize average-policy network $\Pi(s,a\cond \theta^\Pi)$ with random parameters $\theta^\Pi$

    	\State Initialize action-value network $Q(s,a\cond \theta^Q)$ with random parameters $\theta^Q$
    	\State Initialize target network parameters $\theta^{Q'} \gets \theta^Q$
    	\State Initialize anticipatory parameter $\eta$
   	 \ForEach{episode}
   	 	\State Set policy $\sigma \gets \begin{cases} \epsilon\textnormal{-greedy}\left(Q\right), & \mbox{with probability } \eta \\ \Pi, & \mbox{with probability } 1-\eta \end{cases}$
   	 	\State Observe initial information state $s_1$ and reward $r_1$
   	 	\For{$t=1,T$}
   	 	\State Sample action $a_t$ from policy $\sigma$
   	 	\State Execute action $a_t$ in game and observe reward $r_{t+1}$ and next information state $s_{t+1}$
   	 	\State Store transition $\left(s_t,a_t,r_{t+1},s_{t+1}\right)$ in reinforcement learning memory $\mathcal{M}_{RL}$
   	 	\If{agent follows best response policy $\sigma=\epsilon\textnormal{-greedy}\left(Q\right)$}
			\State Store behaviour tuple $\left(s_t,a_t\right)$ in supervised learning memory $\mathcal{M}_{SL}$
   	 	\EndIf
   	 	\State \begin{varwidth}[t]{\linewidth}
   	 		Update $\theta^\Pi$ with stochastic gradient descent on loss\par
   	 		\hskip\algorithmicindent$\mathcal{L}(\theta^\Pi)=\mathbb{E}_{(s,a)\sim\mathcal{M}_{SL}}\left[ -\log\Pi(s,a\cond \theta^\Pi) \right]$
   	 		\end{varwidth}
   	 	\State \begin{varwidth}[t]{\linewidth} 
   	 		Update $\theta^Q$ with stochastic gradient descent on loss\par
   	 		\hskip\algorithmicindent
   	 		$\mathcal{L}\left(\theta^Q\right)=\mathbb{E}_{(s,a,r,s')\sim\mathcal{M}_{RL}}\left[ \left( r+
   	 			\max_{a'}Q(s',a'\cond\theta^{Q'}) - Q(s,a\cond\theta^Q) \right)^2 \right]$
   	 		\end{varwidth}
   	 	\State Periodically update target network parameters $\theta^{Q'} \gets \theta^Q$
   	 	\EndFor
   	\EndFor
   	 	
  	\EndFunction
    \end{algorithmic}
\end{algorithm}

Fictitious play usually keeps track of the average of normal-form best response strategies that players have chosen in the game, $\hat{\pi}_T^i = \frac{1}{T}\sum_{t=1}^T \hat{\beta}_t^i$. \citet{Heinrich15FSP} propose to use sampling and machine learning to generate data on and learn convex combinations of normal-form strategies in extensive form according to equation \eqref{eq:real_equiv}. E.g. we can generate a set of extensive-form data of $\pi_T^i$ by sampling whole episodes of the game, using $\beta_t^i, t=1,...\,,T$, in proportion to their weight, $\frac{1}{T}$, in the convex combination.
\gls{nfsp} agents use reservoir sampling \citep{vitter1985random,osborne2014exponential} to memorize experience of their historical best responses. The agent's supervised learning memory, $\mathcal{M}_{SL}$, is a reservoir to which it only adds experience when following its approximate best response policy, $\beta=\epsilon\textnormal{-greedy}(Q)$.
An \gls{nfsp} agent regularly trains its average policy network, $\Pi(s,a\cond\theta^\Pi)$, to match its average behaviour stored in its supervised learning memory, e.g. by optimizing the log-probability of past actions taken.

If we want all agents to learn simultaneously while playing against each other, we face the following dilemma. 
In principle, each agent could play its average policy, $\pi$, and learn a best response with off-policy Q-learning, i.e. evaluate and maximize its action values, $Q^i(s,a) \approx \mathbb{E}_{\beta^i, \pi^{-i}}\left[G_t^i \Cond S_t = s, A_t = a\right]$, of playing its best response policy $\beta^i$ against its fellow agents' average strategy profile, $\pi^{-i}$. However, in this case the agent would not generate any experience of its own best response behaviour, $\beta$, which is needed to train its average policy network, $\Pi(s,a\cond\theta^\Pi)$, that approximates the agent's average of past best responses.
To address this problem, we suggest using an approximation of \emph{anticipatory dynamics} of continuous-time dynamic fictitious play \citep{shamma2005dynamic}. In this variant of fictitious play players choose best responses to a short-term prediction of their opponents' average normal-form strategies, 
$\hat{\pi}_t^{-i} + \eta \frac{d}{dt} \hat{\pi}_t^{-i}$, where $\eta\in\mathbb{R}$ is termed the \textbf{anticipatory parameter}.
The authors show that for appropriate, game-dependent choice of $\eta$ stability of fictitious play at equilibrium points can be improved. \gls{nfsp} uses $\hat{\beta}_{t+1}^i - \hat{\pi}_{t}^i \approx \frac{d}{dt} \hat{\pi}_t^i$ as a discrete-time approximation of the derivative that is used in these anticipatory dynamics. Note that $\Delta \hat{\pi}_t^i \propto \hat{\beta}_{t+1}^i - \hat{\pi}_{t}^i$ is the normal-form update direction of common discrete-time fictitious play. 
\gls{nfsp} agents choose their actions from the mixture policy $\sigma \equiv (1-\eta)\hat{\pi} + \eta\hat{\beta}$. This enables each agent to compute an approximate best response, $\beta^i$, to its opponents' anticipated average strategy profile, $\sigma^{-i} \equiv \hat{\pi}^{-i}+\eta(\hat{\beta}^{-i}-\hat{\pi}^{-i})$, by iteratively evaluating and maximizing its action values, $Q^i(s,a) \approx \mathbb{E}_{\beta^i, \sigma^{-i}}\left[G_t^i \Cond S_t = s, A_t = a\right]$. Additionally, as each agent's own best response policy is now sampled in proportion to the anticipatory parameter, they can now train their average policy networks from that experience.

\section{Experiments}

We evaluate \gls{nfsp} and related algorithms in Leduc and Limit Texas Hold'em poker games. Most of our experiments measure the exploitability of learned strategy profiles. 
In a two-player zero-sum game, the exploitability of a strategy profile is defined as the expected average payoff that a best response profile achieves against it.
An exploitability of $2\delta$ yields at least a $\delta$-Nash equilibrium. 

\subsection{Leduc Hold'em}

We empirically investigate the convergence of \gls{nfsp} to Nash equilibria in Leduc Hold'em. We also study whether removing or altering some of \gls{nfsp}'s components breaks convergence.

One of our goals is to minimize reliance on prior knowledge. Therefore, we attempt to define a domain-independent encoding of information states in poker games. Contrary to other work on computer poker \citep{zinkevich2007regret,gilpin2007gradient,johanson2013evaluating}, we do not engineer any higher-level features.
Poker games usually consist of multiple rounds. At each round new cards are revealed to the players. We represent each rounds' cards by a k-of-n encoding. E.g. \gls{lhe} has a card deck of $52$ cards and on the second round three cards are revealed. Thus, this round is encoded with a vector of length $52$ and three elements set to $1$ and the rest to $0$. In Limit Hold'em poker games, players usually have three actions to choose from, namely \{fold, call, raise\}. Note that depending on context, calls and raises can be referred to as checks and bets respectively. Betting is capped at a fixed number of raises per round. Thus, we can represent the betting history as a tensor with $4$ dimensions, namely \{player, round, number of raises, action taken\}. E.g. heads-up \gls{lhe} contains $2$ players, $4$ rounds, $0$ to $4$ raises per round and $3$ actions. Thus we can represent a \gls{lhe} betting history as a $2\times4\times5\times3$ tensor. In a heads-up game we do not need to encode the fold action, as a two-player game always ends if one player gives up. Thus, we can flatten the $4$-dimensional tensor to a vector of length $80$. Concatenating with the card inputs of $4$ rounds, we encode an information state of \gls{lhe} as a vector of length $288$. Similarly, an information state of Leduc Hold'em can be encoded as a vector of length $30$, as it contains $6$ cards with 3 duplicates, $2$ rounds, $0$ to $2$ raises per round and $3$ actions.

For learning in Leduc Hold'em,
we manually calibrated \gls{nfsp} for a fully connected neural network with $1$ hidden layer of $64$ neurons and rectified linear activations. We then repeated the experiment for various network architectures with the same parameters. In particular, we set the sizes of memories to $200$k and $2$m for $\mathcal{M}_{RL}$ and $\mathcal{M}_{SL}$ respectively. $\mathcal{M}_{RL}$ functioned as a circular buffer containing a recent window of experience. $\mathcal{M}_{SL}$ was updated with reservoir sampling \citep{vitter1985random}. The reinforcement and supervised learning rates were set to $0.1$ and $0.005$, and both used vanilla \gls{sgd} without momentum for stochastic optimization of the neural networks. Each agent performed $2$ stochastic gradient updates of mini-batch size $128$ per network for every $128$ steps in the game. The target network of the \gls{dqn} algorithm was refitted every $300$ updates. \gls{nfsp}'s anticipatory parameter was set to $\eta=0.1$. The $\epsilon$-greedy policies' exploration started at $0.06$ and decayed to $0$, proportionally to the inverse square root of the number of iterations.

\begin{figure}
		\centering
    \subfloat[Varying network sizes\label{fig:leduc_nets}]{%
      \includegraphics[width=0.32\textwidth]{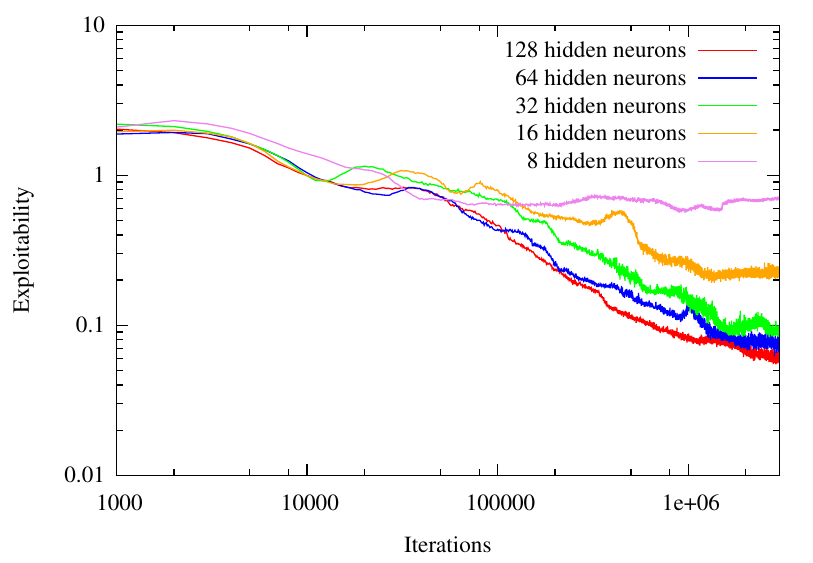}
    }
    \hfill
    \subfloat[Removed essential components\label{fig:leduc_fails}]{%
      \includegraphics[width=0.32\textwidth]{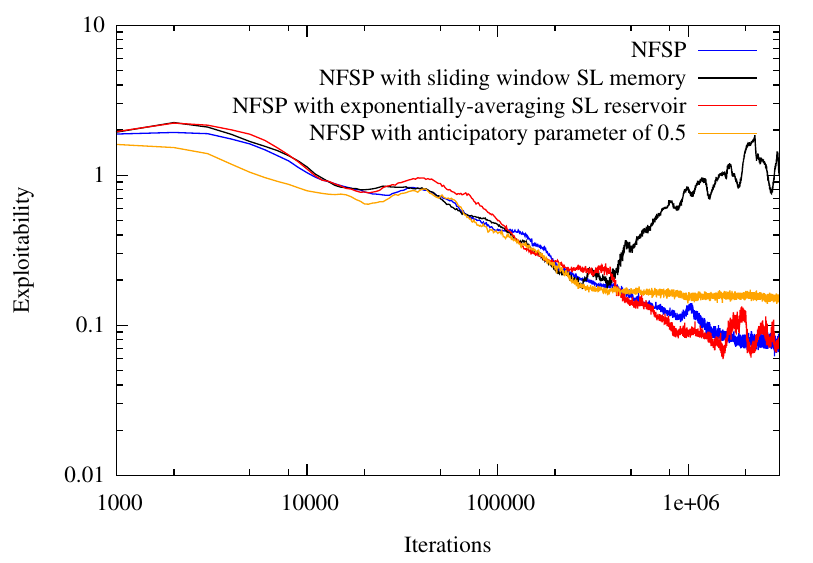}
    }
    \hfill
    \subfloat[Comparison to \gls{dqn}\label{fig:dqn}]{%
      \includegraphics[width=0.32\textwidth]{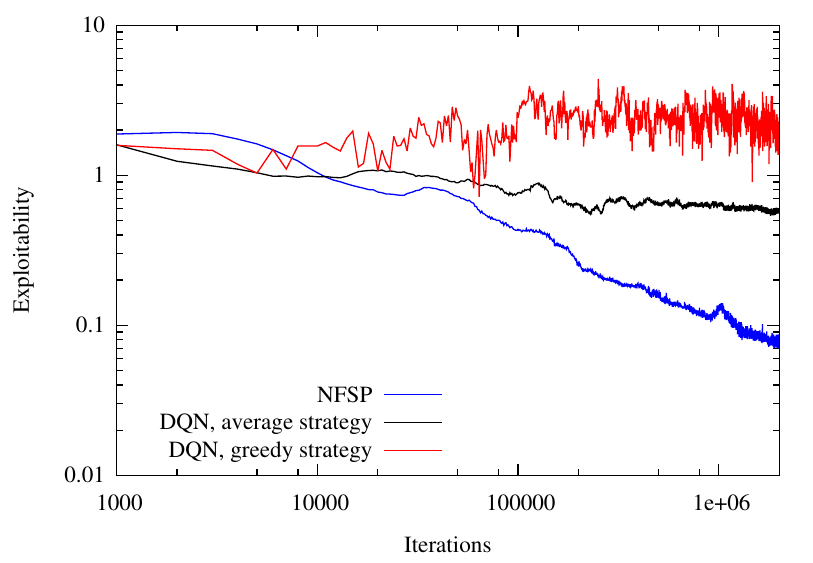}
    }
    \caption{Learning performance of \gls{nfsp} in Leduc Hold'em.}
    \label{fig:nfsp_leduc}
\end{figure}

Figure \ref{fig:leduc_nets} shows \gls{nfsp} approaching Nash equilibria for various network architectures. We observe a monotonic performance increase with size of the networks. 
\gls{nfsp} achieved an exploitability of $0.06$, which full-width \gls{xfp} typically achieves after around $1000$ full-width iterations.

In order to investigate the relevance of various components of \gls{nfsp}, e.g. reservoir sampling and anticipatory dynamics, we conducted an experiment that isolated their effects.
Figure \ref{fig:leduc_fails} shows that these modifications led to decremental performance. In particular, using a fixed-size sliding window to store experience of the agents' own behaviour led to divergence. \gls{nfsp}'s performance plateaued for a high anticipatory parameter of $0.5$, that most likely violates the game-dependent stability conditions of dynamic fictitious play \citep{shamma2005dynamic}. 
Finally, using exponentially-averaged reservoir sampling for supervised learning memory updates led to noisy performance.

\subsection{Comparison to \gls{dqn}}

Several stable algorithms have previously been proposed for deep reinforcement learning, notably the \gls{dqn} algorithm \citep{mnih2015human}. However, the empirical stability of these algorithms was only previously established in single-agent, perfect (or near-perfect) information \glspl{mdp}. Here, we investigate the stability of \gls{dqn} in multi-agent, imperfect-information games, in comparison to \gls{nfsp}.

\gls{dqn} learns a deterministic, greedy strategy. 
Such strategies are sufficient to behave optimally in single-agent domains, i.e. \glspl{mdp} for which \gls{dqn} was designed.
However, imperfect-information games generally require stochastic strategies to achieve optimal behaviour.
One might wonder if the average behaviour of \gls{dqn} converges to a Nash equilibrium. To test this, we augmented \gls{dqn} with a supervised learning memory and train a neural network to estimate its average strategy. Unlike \gls{nfsp}, the average strategy does not affect the agent's behaviour in any way; it is passively observing the \gls{dqn} agent to estimate its evolution over time. We implement this variant of \gls{dqn} by using \gls{nfsp} with an anticipatory parameter of $\eta=1$. We trained \gls{dqn} with all combinations of the following parameters: Learning rate $\{0.2,0.1,0.05\}$, decaying exploration starting at $\{0.06, 0.12\}$ and reinforcement learning memory $\{ 2\textnormal{m reservoir},2\textnormal{m sliding window} \}$.
Other parameters were fixed to the same values as \gls{nfsp}; note that these parameters only affect the passive observation process.
We then chose the best-performing result of \gls{dqn} and compared to \gls{nfsp}'s performance that was achieved in the previous section's experiment.
\gls{dqn} achieved its best-performing result with a learning rate of $0.1$, exploration starting at $0.12$ and a sliding window memory of size $2$m.

Figure \ref{fig:dqn} shows that \gls{dqn}'s deterministic strategy is highly exploitable, which is expected as imperfect-information games usually require stochastic policies. \gls{dqn}'s average behaviour does not approach a Nash equilibrium either. 
In principle, \gls{dqn} also learns a best-response to the historical experience generated by fellow agents. So why does it perform worse than \gls{nfsp}? 
The problem is that \gls{dqn} agents exclusively generate  
self-play experience according to their $\epsilon$-greedy strategies. 
These experiences are both highly correlated over time, and highly focused on a narrow distribution of states. In contrast, \gls{nfsp} agents use an ever more slowly changing (anticipated) average policy to generate self-play experience. 
Thus, their experience varies more smoothly, resulting in a more stable data distribution, and therefore more stable neural networks. Note that this issue is not limited to \gls{dqn}; other common reinforcement learning methods have been shown to exhibit similarly stagnating performance in poker games \citep{heinrich2015smooth}.

\subsection{Limit Texas Hold'em}

We applied \gls{nfsp} to \gls{lhe}, a game that is popular with humans.
Since in 2008 a computer program beat expert human \gls{lhe} players for the first time in a public competition,
modern computer agents are widely considered to have achieved superhuman performance \citep{newall2013further}. The game was \emph{essentially solved} by \citet{bowling2015heads}. 
We evaluated our agents against the top 3 computer programs of the most recent (2014) \gls{acpc} that featured \gls{lhe}.
Learning performance was measured in milli-big-blinds won per hand, mbb/h, i.e. one thousandth of a big blind that  players post at the beginning of a hand.

\begin{figure}
	\centering
    \subfloat[Win rates against SmooCT. The estimated standard error of each evaluation is less than $10$ mbb/h.\label{fig:holdem}]{
    \begin{minipage}[c][0.75\width]{0.45\textwidth}
      \includegraphics[width=\textwidth]{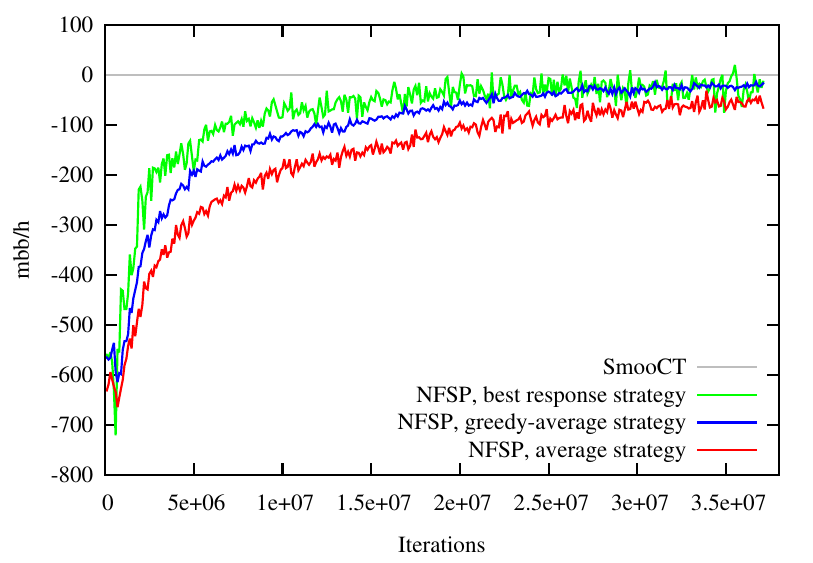}
    \end{minipage}
    }
    \hfill
    \subfloat[Win rates of \gls{nfsp}'s greedy-average strategy against the top 3 agents of the \gls{acpc} 2014.\label{tab:huhu}]{
    		\begin{minipage}[c][0.75\width]{0.45\textwidth}
			  \begin{tabular}[b]{| l  l |}
			    \hline
			     Match-up & Win rate (mbb/h)  \\ 
			    \hline
			    escabeche & -52.1 $\pm$  8.5 \\ 
			    SmooCT & -17.4 $\pm$  9.0  \\ 
			    Hyperborean & -13.6 $\pm$   9.2 \\ \hline
			  \end{tabular}
			\end{minipage}
    }
    \caption{Performance of \gls{nfsp} in Limit Texas Hold'em.}
    \label{fig:holdem_results}
  \end{figure}

We manually calibrated \gls{nfsp} by trying $9$ configurations.
We achieved the best performance with the following parameters. 
The neural networks were fully connected with four hidden layers of $1024,512,1024$ and $512$ neurons with rectified linear activations.
The memory sizes were set to $600$k and $30$m for $\mathcal{M}_{RL}$ and $\mathcal{M}_{SL}$ respectively. $\mathcal{M}_{RL}$ functioned as a circular buffer containing a recent window of experience. $\mathcal{M}_{SL}$ was updated with exponentially-averaged reservoir sampling \citep{osborne2014exponential}, replacing entries in $\mathcal{M}_{SL}$ with minimum probability $0.25$.
We used vanilla \gls{sgd} without momentum for both reinforcement and supervised learning, with learning rates set to $0.1$ and $0.01$ respectively. Each agent performed $2$ stochastic gradient updates of mini-batch size $256$ per network for every $256$ steps in the game. 
The target network 
was refitted every $1000$ updates.
\gls{nfsp}'s anticipatory parameter was set to $\eta=0.1$. The $\epsilon$-greedy policies' exploration started at $0.08$ and decayed to $0$, more slowly than in Leduc Hold'em. In addition to \gls{nfsp}'s main, average strategy profile we also evaluated the best response and greedy-average strategies, which deterministically choose actions that maximize the predicted action values or probabilities respectively.

To provide some intuition for win rates in heads-up \gls{lhe}, a player that always folds will lose 750 mbb/h, and expert human players typically achieve expected win rates of 40-60 mbb/h at online high-stakes games.
Similarly, the top half of computer agents in the \gls{acpc} 2014 achieved up to 50 mbb/h between themselves.
While training, we periodically evaluated \gls{nfsp}'s performance against SmooCT from symmetric play for $25000$ hands each. Figure \ref{fig:holdem} presents the learning performance of \gls{nfsp}. 
\gls{nfsp}'s average and greedy-average strategy profiles exhibit a stable and relatively monotonic performance improvement, and achieve win rates of around -$50$ and -$20$ mbb/h respectively. The best response strategy profile exhibited more noisy performance, mostly ranging between -$50$ and $0$ mbb/h. We also evaluated the final greedy-average strategy against the other top 3 competitors of the \gls{acpc} 2014. Table \ref{tab:huhu} presents the results. \gls{nfsp} achieves winrates similar to those of the top half of computer agents in the \gls{acpc} 2014 and thus is competitive with superhuman computer poker programs.

\subsection{Approximations to fictitious play}
NFSP approximates fictitious play, both by using a neural network function approximator to represent strategies, and by averaging those strategies via gradient descent machine learning. In the appendix we provide experiments that illustrate the effect of these two approximations, compared to exact representations of strategies and perfect averaging procedures.

\section{Related work}

Reliance on human expert knowledge can be expensive, prone to human biases and limiting if such knowledge is suboptimal. Yet many methods that have been applied to games have relied on human expert knowledge. 
Deep Blue used a human-engineered evaluation function for chess \citep{campbell2002deep}.
In computer Go, \citet{maddison2014move} and \citet{clark2015training} trained deep neural networks from data of expert human play.
In computer poker, current game-theoretic approaches use heuristics of card strength to abstract the game to a tractable size \citep{zinkevich2007regret,gilpin2007gradient,johanson2013evaluating}. 
\citet{regress15} recently combined one of these methods with function approximation. However, their full-width algorithm has to implicitly reason about all information states at each iteration, which is prohibitively expensive in large domains.
In contrast, NFSP focuses on the sample-based reinforcement learning setting where the game's states need not be exhaustively enumerated and the learner may not even have a model of the game's dynamics.

Nash equilibria are the only strategy profiles that rational agents can hope to converge on in self-play \citep{bowling2001rational}. 
TD-Gammon \citep{tesauro1995temporal} is a world-class backgammon agent, whose main component is a neural network trained from self-play reinforcement learning. 
While its algorithm, based on temporal-difference learning, is sound in two-player zero-sum perfect-information games, it does not generally converge in games with imperfect information.
\gls{dqn} \citep{mnih2015human} combines temporal-difference learning with experience replay and deep neural network function approximation. It achieved human-level performance in a majority of Atari games, learning from raw sensory inputs. However, these Atari games were set up as single-agent domains with potential opponents fixed and controlled by the Atari emulator. Our experiments showed that \gls{dqn} agents were unable to approach a Nash equilibrium in Leduc Hold'em, where players were allowed to adapt dynamically.
\citet{yakovenko2016poker} trained deep neural networks in self-play in computer poker, including two poker games that are popular with humans. 
Their networks performed strongly against heuristic-based and simple computer programs. Expert human players were able to outperform their agent, albeit over a statistically insignificant sample size. 
It remains to be seen whether their approach converges in practice or theory.

In this work, we focused on imperfect-information two-player zero-sum games. Fictitious play, however, is also guaranteed to converge to Nash equilibria in cooperative, potential games \citep{monderer1996fictitious}. It is therefore conceivable that \gls{nfsp} can be successfully applied to these games as well.
Furthermore, recent developments in continuous-action reinforcement learning \citep{lillicrap2015continuous} could enable \gls{nfsp} to be applied to continuous-action games, which current game-theoretic methods cannot deal with directly.

\section{Conclusion}

We have introduced \gls{nfsp}, the first end-to-end deep reinforcement learning approach to learning approximate Nash equilibria of imperfect-information games from self-play.
Unlike previous game-theoretic methods, \gls{nfsp} is scalable without prior domain knowledge. Furthermore, \gls{nfsp} is the first deep reinforcement learning method known to converge to approximate Nash equilibria in self-play.
Our experiments have shown \gls{nfsp} to converge reliably to approximate Nash equilibria in a small poker game, whereas \gls{dqn}'s greedy and average strategies did not.
\gls{nfsp} learned a strategy that is competitive with superhuman programs in a real-world scale imperfect-information game from scratch without using explicit prior knowledge.

\subsubsection*{Acknowledgments}
We thank Peter Dayan, Marc Lanctot and Marc Bellemare for helpful discussions and feedback.
This research was supported by the UK Centre for Doctoral Training in Financial Computing and by the NVIDIA Corporation.

%

{
\bibliographystyle{apalike}
\small
\bibliography{nfsp}
}

\appendix

\section{Robustness of XFP}

To understand how function approximation interacts with \gls{fsp}, we conducted some simple experiments 
that emulate approximation and sampling errors in the full-width algorithm \gls{xfp}. 
Firstly, we explore what happens when the perfect averaging used in XFP is replaced by an incremental averaging process closer to gradient descent. 
Secondly, we explore what happens when the exact table lookup used in XFP is replaced by an approximation with epsilon error.

\begin{figure}[!ht]
    \subfloat[Constant stepsizes in average policy updates\label{fig:xfp_lr}]{%
      \includegraphics[width=0.45\textwidth]{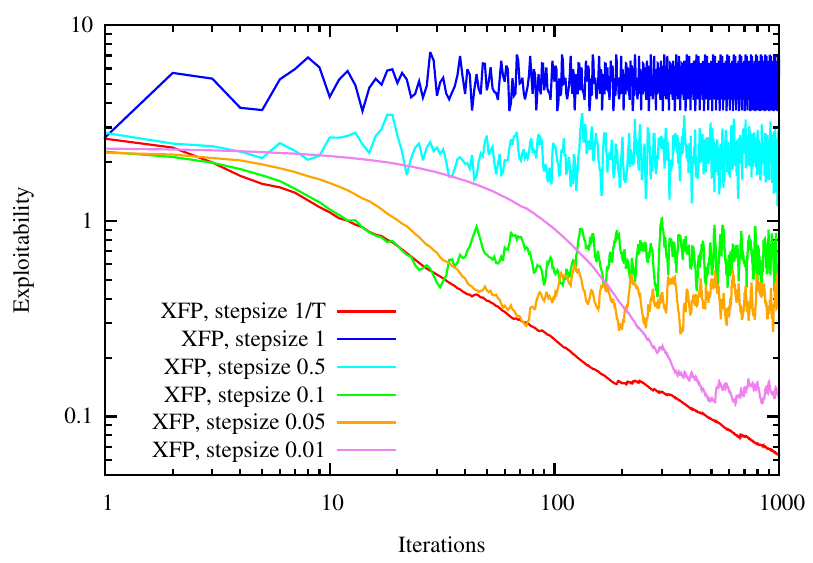}
    }
    \hfill
    \subfloat[Uniform-random noise in best response computation\label{fig:xfp_epsbr}]{%
      \includegraphics[width=0.45\textwidth]{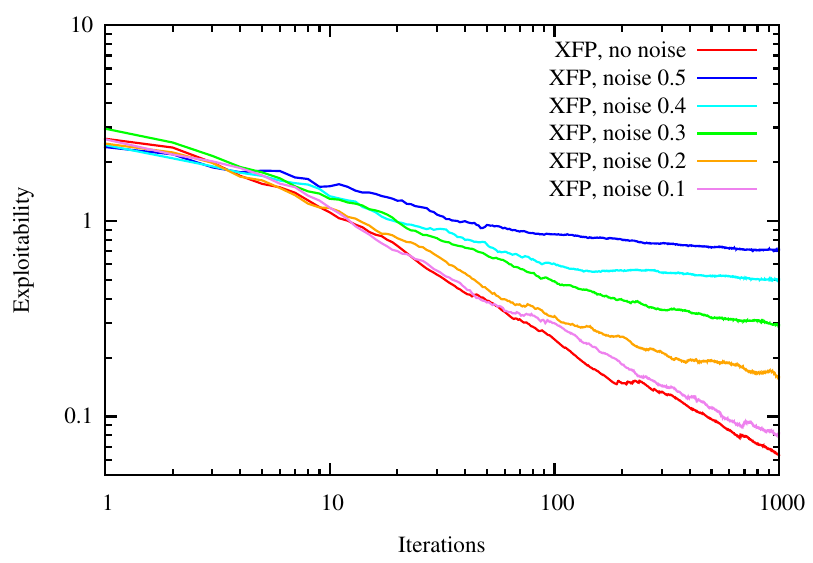}
    }
    \caption{Full-width \gls{xfp} performance with added noise}
    \label{fig:xfp}
  \end{figure}

Figure \ref{fig:xfp_lr} shows the performance of \gls{xfp} with default, $1/T$, and constant stepsizes for its strategy updates.  
We see improved asymptotic but lower initial performance for smaller stepsizes. For constant stepsizes the performance seems to plateau 
rather than diverge.
With reservoir sampling we can achieve an effective stepsize of $1/T$. However, the results suggest that exponentially-averaged reservoir sampling can be a viable choice too,
as exponential averaging of past memories would approximately correspond to using a constant stepsize.

\gls{xfp} with stepsize $1$ is equivalent to a full-width iterated best response algorithm. While this algorithm converges to a Nash equilibrium in finite perfect-information two-player zero-sum games, the results suggest that with imperfect information this is not generally the case. The Poker-CNN algorithm introduced by
\citet{yakovenko2016poker} stores a small number of past strategies which it iteratively computes new strategies against. Replacing strategies in that set is similar to updating an average strategy with a large stepsize. This might lead to similar problems as shown in Figure \ref{fig:xfp_lr}.

Our \gls{nfsp} agents add random exploration to their policies and use noisy stochastic gradient updates to learn action values, which determine their approximate best responses. 
Therefore, we investigated the impact of random noise added to the best response computation, which \gls{xfp} performs by dynamic programming. At each backward induction step, we pass back a uniform-random action's value with probability $\epsilon$ and the best action's value otherwise.
Figure \ref{fig:xfp_epsbr} shows monotonically decreasing performance with added noise. However, performance remains stable and keeps improving for all noise levels. 

\end{document}